\documentclass{article}

\usepackage{spconf,amsmath,graphicx}

\usepackage{amssymb}
\usepackage{dsfont}
\usepackage{amsfonts}
\usepackage{mathrsfs}

\usepackage{setspace}
\usepackage{xspace}

\usepackage{ stmaryrd }
\usepackage{lineno}
\usepackage{color}
\usepackage{url}
\usepackage{ stmaryrd }

\usepackage{amssymb,color} 
\usepackage{algorithm}
\usepackage{algpseudocode}

\usepackage{soul}

\newcommand{\Rr}{\mathds{R}}
\newcommand{\Cr}{\mathds{C}}

\newcommand{\R}{P}

\newcommand{\Q}{Q}

\newcommand{\XI}{{X}}

\newcommand{\AAA}{\mathbf{X}} 
\newcommand{\BBB}{\mathbf{Y}}

\newcommand{\eg}{\textit{e.g.}, }
\newcommand{\ie}{\textit{i.e.}, }

\DeclareMathOperator*{\argmin}{arg\,min}

\newtheorem{theorem}{Theorem}
\newtheorem{remark}{Remark}

\title{Optimal Low-Rank Dynamic Mode Decomposition}

\name{Patrick H\'eas and C\'edric Herzet}
\address{INRIA Centre Rennes - Bretagne Atlantique, Campus universitaire de Beaulieu, 35000 Rennes, France}
\begin{document}
%
\maketitle
\begin{abstract}
Dynamic Mode Decomposition (DMD) has  emerged as a powerful tool for analyzing  the dynamics of non-linear systems from experimental datasets. Recently, several attempts have extended   DMD to the context of low-rank approximations. This extension is of particular interest for reduced-order modeling in various applicative domains, \eg for climate prediction, to study  molecular dynamics or micro-electromechanical devices.    This low-rank  extension takes the form of a non-convex optimization problem. 
To the best of our knowledge, only  sub-optimal algorithms have been proposed in the literature to compute the solution of this problem. In this paper, we prove that there exists a closed-form optimal solution to this problem and design an effective algorithm to compute it based  on Singular Value Decomposition (SVD). A toy-example illustrates the gain in performance  of  the proposed algorithm compared to  state-of-the-art techniques.
\end{abstract}
\begin{keywords}
Low-Rank Approximations, Reduced-Order Models, Dynamical Mode Decomposition, SVD
\end{keywords}

\section{Introduction}

In many fields of Sciences, one is interested in studying the {spatio-temporal} evolution of a state variable characterized by  a  partial differential equation.  Numerical discretization in space and time leads to a high dimensional system of equations {of the form:} \vspace{-0.cm}
\begin{align}\label{eq:model_init} 
 \left\{\begin{aligned}
& x_{t}= f_t(x_{t-1}) , \\
&x_1={\theta},
\end{aligned}\right. \vspace{-0.cm}
\end{align} 
\noindent
 {where} each element   of the sequence of state variables  $\{x_t\}_t$ belongs to $\Rr^n$,   $f_t:\Rr^n \to \Rr^n$ with the initial condition  $\theta \in \Rr^n$. 
 {Because \eqref{eq:model_init} may correspond to a very high-dimensional system in some applications, computing a trajectory $\{x_t\}_t$ given an initial condition ${\theta}$}  may  lead to a heavy computational load, which may prohibit the direct use of the original high-dimensional system.  
 
The context of uncertainty quantification provides an appealing example. Assume we are interested in characterizing the distribution of  random trajectories generated by \eqref{eq:model_init} with respect to the distribution of the initial condition. A straightforward approach would be to sample the initial condition and run the high-dimensional system. However, in many applicative contexts, it is  impossible to generate enough trajectories  to make accurate approximations with Monte-Carlo techniques.

 As a response to this computational bottleneck,  reduced-order models aim to approximate  the trajectories of the system for a range of regimes determined by a set of initial conditions \cite{2015arXiv150206797C}. 
 A common approach is to assume that the trajectories of interest are well approximated in a sub-space of $\Rr^n$. In this spirit, many  tractable low-rank approximations of  high-dimensional systems have been proposed in the literature, the most familiar  being  proper orthogonal decomposition (POD)~\cite{9780511622700}, balanced truncation~\cite{Antoulas2005Overview}, Taylor expansions~\cite{ZAMM:ZAMM19830630105} or reduced-basis techniques~\cite{Quarteroni2011Certified}.  Other popular sub-space methods,  such as  linear inverse modeling (LIM)~\cite{penland1993prediction}, principal oscillating patterns (POP)~\cite{Hasselmann88}, or more recently, dynamic mode decomposition (DMD)~\cite{Schmid10,Chen12,Jovanovic12,williams2015data,Tu2014391}, are known as Koopman operator approximations.  
 
In this paper, we  consider the setting where  system \eqref{eq:model_init} is a black-box. In other words, we assume that  we do not know the exact form of $f_t$ in \eqref{eq:model_init} and we only have access to a set of representative trajectories $\{x^i_t\}_{t,i}$, $i=1,...,N$, $t=1,...,T$ so-called \textit{snapshots}, obtained by running the high-dimensional system for $N$ different initial conditions.   
Moreover, we focus on the low-rank DMD approximation problem studied  in \cite{Chen12,Jovanovic12}.  In a nutshell, these studies provide a procedure for determining  a matrix $ \hat A_k{\in\Rr^{n\times n}}$ of rank $k \ll n$, which substitutes {for} function $f_t$ in \eqref{eq:model_init} {as}
\begin{align}\label{eq:model_koopman_approx} 
 \left\{\begin{aligned}
& \tilde x_{t}=  \hat A_k \tilde  x_{t-1} , \\
&\tilde x_1={\theta},
\end{aligned}\right. \vspace{-0.cm}
\end{align} 
and generates the approximations $\tilde x_{t}\in \Rr^n$ with a low computational effort.  Alternatively, given  $\hat A_k$, and its  $k$ non-zero eigenvalues $\lambda_i \in \Cr,\, i=1 \cdots k$  and associated eigenvectors $\phi_i \in \Cr^n,\, i=1 \cdots k$,  trajectories of  \eqref{eq:model_koopman_approx} can be computed by using the reduced-order model   
\begin{align}
\tilde x_{t}=\sum_{i=1}^k \nu_{i,t} \phi_i,\label{eq:koopman1}\quad \nu_{i,t} =  \lambda_i^{t-1} \phi_i^*\theta,
\end{align} 
as long as matrix $\hat A_k$ is symmetric. We will assume it is always the case for simplification issues.
In what follows, we will  refer to the parameters  $\phi_i$ and $\nu_{i,t}\in \Cr$  as  the  \textit{$i$-th low-rank DMD mode and amplitude} at time $t$. 

Matrix $\hat A_k$ targets the solution of the following non-convex optimization problem, which we will refer to as the {\it low-rank DMD approximation problem}
\begin{align}\label{eq:prob} 
A_k^\star \in &\argmin_{A:\textrm{rank}(A)\le k}  \sum_{t,i} \| x^i_{t} - A   x^i_{t-1}\|^2_2, 
\end{align} 
where $\|\cdot\|_2$ refers to the $\ell_2$ norm.  
In order to compute {a} solution $A_k^\star$, the authors in \cite{Chen12,Jovanovic12} propose to rely on the  assumption  of linear dependence of recent snapshots on previous ones. This assumption may  not be reasonable, especially  in the case of non-linear systems.

Beyond the reduced modeling context discussed above,   there has been a resurgence of interest for  low-rank solutions of linear matrix equations \cite{recht2010guaranteed}.  This  class of  problems is very large and includes in particular  problem \eqref{eq:prob}. Problems in this class  are generally nonconvex and do not  admit explicit solutions.  Howewer,  important results have arisen at the theoretical  and  algorithmic level, enabling the characterization of the solution for this class of problems by convex relaxation \cite{fazel2002matrix}. Applications concern scenarios such as  low-rank matrix completion, image compression or  minimum order linear system realization, see  \cite{recht2010guaranteed}.  Nevertheless, there exists  certain instances with a very special structure, which admit closed-form solutions \cite{parrilo2000cone,mesbahi1997rank}.  This occurs typically when the solution can be deduced from the  well-known Eckart-Young theorem \cite{eckart1936approximation}. 
 

The contribution of this paper is to show that the special structure of problem \eqref{eq:prob} enables the characterization of  an exact closed-form solution  and  an easily implementable solver based on singular value decomposition (SVD). In the case  $k \ge N(T-1)$, the proposed algorithm  computes the solution of \cite{Tu2014391}. 
More interestingly, for $k < N(T-1)$, \ie in the constrained case, our approach enables to solve exactly the low-rank DMD approximation problem  without 
 1) any  assumption of linear dependence,
 2) the use of an iterative solver, on the contrary to the approaches proposed in \cite{Chen12,Jovanovic12,fazel2002matrix}. 

The paper is organized as follows. In section \ref{sec:stateArt}, we provide a brief review of state-of-the-art techniques to compute low-rank DMD of experimental data. Section \ref{sec:contrib}  details our analytical solution and the algorithm solving \eqref{eq:prob}.  Given this optimal solution, it then presents the  reduced-order model solving  \eqref{eq:model_koopman_approx}. Finally,  a numerical evaluation of the method is presented in {Section}  \ref{sec:numEval} and  concluding remarks are {given} in a last section.

\section{State-Of-The-Art Overview}\label{sec:stateArt}

In what follows,  we assume that we have at our disposal  $N$ trajectories of $T$ snapshots.    We will  need in the following some matrix notations. The symbol $\|\cdot\|_F$ and the upper script $\cdot^*$ will respectively refer to the Frobenius norm and the transpose operator. $I_k$ will denote the $k$-dimensional identity matrix. 
 Let consecutive elements of the $i$-th snapshot trajectory between time $t_1$ and $t_2$  be  gathered in a matrix   $\XI_{t_1:t_2}^i = (x^i_{t_1},\cdots,x^i_{t_2}),$ and let  two large matrices  $ \AAA, \BBB \in \Rr^{n \times  m} $ with $m=(T-1)N$ be defined as 
  \begin{align}\label{eq:matrixAB}
 \AAA = (\XI^1_{1:T-1}, ..., \XI^N_{1:T-1}), \quad 
 \BBB= ( \XI^1_{2:T}, ...,  \XI^N_{2:T}) .
 \end{align}
 Without loss of generality, this work will assume that $m\leq n$ and that $\textrm{rank}(\AAA)=\textrm{rank}(\BBB)=m$.  
 We introduce the SVD decomposition of a matrix $M\in \Rr^{p \times q }$ with $p\ge q$: $M=W_M\Sigma_M V_M^*$ with $W_M\in \Rr^{p \times q }$, $V_M\in \Rr^{ q \times  q}$ and $\Sigma_M\in \Rr^{q  \times q }$ so that $W_M^*W_M=V_M^*V_M=I_q$ and $\Sigma_M$ is diagonal. The  Moore-Penrose pseudo-inverse of a matrix  $M$  will be defined as 
 $M^\dagger=V_M\Sigma_M^{-1} W_M^*$.
 
With these notations, problem \eqref{eq:prob} can be rewritten as 
		\begin{align}\label{eq:prob1} 
		A_k^\star \in &\argmin_{A:\textrm{rank}(A)\le k} \|\BBB -A \AAA \|_F^2.
		\end{align} 

In what follows, we begin by presenting two state-of-the-art methods which enable to compute an approximation of the solution of problem  \eqref{eq:prob1}.
 
 \subsection{Projected DMD and Low-Rank Formulation}
As detailed herafter, the original DMD approach  first proposed in \cite{Schmid10}, so-called \textit{projected} DMD in \cite{Tu2014391}, assumes that columns of $A\AAA$ are in the span of $\AAA $. The assumption is written by the authors in  \cite{Schmid10,Jovanovic12} as the existence of $A^c\in \Rr^{ m \times m}$, the so-called  \textit{companion matrix} of $A$ parametrized by $m$ coefficients, such that
\begin{align}
A \AAA=\AAA  A^c.\label{eq:companion}
  \end{align} 
  We remark that this assumption is in particular valid when the $i$-th snapshot $x^i_T$ can be expressed as a linear combination of the columns of $\XI_{1:T-1}^i $ and when $f_t$ is linear. 
  Using the  SVD decomposition $\AAA=W_\AAA\Sigma_\AAA V_\AAA^*$ and noticing $\AAA$ is full rank, we obtain  from \eqref{eq:companion}  a projected representation of $A$ in the basis spanned by the columns of $W_\AAA$,  
\begin{align}\label{eq:DMDassumption}
 W_\AAA^*AW_\AAA=\tilde A^c,
\end{align}
where $ \tilde A^c=\Sigma_\AAA V_\AAA^*A^cV_\AAA\Sigma_\AAA^{-1}\in \Rr^{ m \times  m}.$ 
Therefore, the low-rank  formulation in \cite{Jovanovic12} proposes to approach the solution of \eqref{eq:prob1} by determining the $m$ coefficients of matrix $A^c$ which minimize  the Frobenius norm of   the residual $ \BBB -A \AAA $. 
This yields after some algebraic manipulations to solve the problem 
\begin{align}\label{eq:DMDSVD}
\argmin_{ \tilde A^c: \textrm{rank}(\tilde A^c\Sigma_\AAA)\le k} \|W_\AAA^*\BBB V_\AAA - \tilde A^c \Sigma_\AAA\|^2_F.
\end{align}
 The Eckart-Young theorem \cite{eckart1936approximation} then provides the optimal  solution  to this problem based  on a  rank-$k$  SVD approximation of matrix $B=W_\AAA^*\BBB V_\AAA$ given by  $W_B\Lambda_BV_B^*$ where $\Lambda_B$ is a diagonal matrix containing only the $k$-largest singular values of $\Sigma_B$ and with zero entries {otherwise}. 
  Exploiting the  low-dimensional representation \eqref{eq:DMDassumption}, a reduced-order model for trajectories can then be obtained  by inserting in \eqref{eq:model_koopman_approx} the low-rank approximation  
 \begin{align}\label{eq:projDMD}
 \hat A_k=W_\AAA W_B\Lambda_BV_B^*\Sigma_\AAA^{-1}W_\AAA^*.
 \end{align}
   As an alternative, the authors propose a reduced-order model for trajectories relying on the so-called DMD modes and their amplitudes. These modes are related to the eigenvectors of the solution of \eqref{eq:DMDSVD}.  The amplitudes are  given by solving a convex optimization problem with an iterative gradient-based method, see details in \cite{Jovanovic12}.  
 
\subsection{Non-projected DMD}
If we remove the low-rank constraint, 
  \eqref{eq:prob1} becomes a least-squares  problem  
whose solution is
\begin{align}\label{eq:exactDMD}
\hat A_m=\BBB\AAA^{\dagger}=\BBB V_\AAA\Sigma_\AAA^{-1}W_\AAA^*.
\end{align}
 Based on the approximation $\hat A_m$, DMD modes and amplitudes serve to  design a   model to reconstruct  trajectories of \eqref{eq:model_koopman_approx}.
 We note that the DMD modes  are simply given by the eigendecomposition of $\hat A_m$, which can be  {efficiently} 
 {computed} using SVD, 
  as proposed in \cite{Tu2014391}. The {associated} DMD amplitudes can then  easily be derived.  

 It is important to remark that  truncating  to a rank-$k$ the solution of the above  unconstrained minimization problem  will not necessarily yield the solution of \eqref{eq:prob1}. This approach   will generally be  sub-optimal. 
However surprisingly, the solution to problem \eqref{eq:prob1}  remain to our knowledge  overlooked in the literature, and no algorithms enabling  non-projected low-rank DMD approximations have yet been proposed.

\section{ The Proposed Approach}\label{sec:contrib}

\subsection{Closed-form Solution to \eqref{eq:prob1}}
  Let the columns of matrix  $\R \in \Rr^{n \times k}$  be the real orthonormal  eigenvectors  associated  to the  $k$ largest  eigenvalues  of    matrix  $\BBB\BBB^*.$    
\begin{theorem}\label{prop22}
A solution of \eqref{eq:prob1}  is $A_k^\star=\R  \R^*\BBB \AAA^{\dagger}.$

 \end{theorem}
This theorem states that \eqref{eq:prob1} can be  simply solved by computing the orthogonal projection of the    unconstrained problem solution \eqref{eq:exactDMD}  onto the subspace spanned by the $k$ first eigenvectors of $\BBB \BBB^*$. A detailed proof is provided in {the technical report associated to this paper \cite{Heas16_DMD}.}

\subsection{{Efficient} Solver}

The matrix $\BBB\BBB^*$ is  of size $n\times n$.   {Since $n$ is typically very large, t}his prohibits the direct computation of an eigenvalue decomposition. 
The following well-know remark is useful to overcome this difficulty.  \\

\begin{remark}\label{rem:1}
The   eigenvectors  associated to the $m~\le~n$  non-zero eigenvalues of  matrix  $\BBB\BBB^*\in \Rr^{n \times n}$ with   $ \BBB \in \Rr^{n \times m}$ can be obtained {from} 
  the eigenvectors $ V_\BBB=(v_1,...,v_{m}) \in \Rr^{m \times m}$
 and eigenvalues  
  of the smaller  matrix $\BBB^*\BBB \in \Rr^{m\times m}$. Indeed,
the SVD  of   a matrix  $ \BBB$ of rank  $m$ is 
$\BBB=  W_\BBB\Sigma_\BBB    V_\BBB^*,$
 where the 
  columns 
  of  matrix  $  W_\BBB \in \Rr^{n \times m} $ are the eigenvectors of $\BBB\BBB^*$. 
  Since $  V_\BBB$ is unitary, we obtain that the sought vectors are the {first $k$} 
  columns of  $ W_\BBB$, \ie of $ \BBB  V_\BBB \,\Sigma_\BBB^{-1}.$
  \end{remark}

In the light of  this remark, it is straightforward to design Algorithm \ref{algo:1}, which will      compute efficiently  the solution  of \eqref{eq:prob1}  based on  SVDs.

\begin{algorithm}[t]
\begin{algorithmic}[0]
\State \textbf{input}: $N$-sample $\{\XI_{1:T}^i\}_{i=1}^N$
\State 1) Form matrix $\AAA$ and $\BBB$ as defined in \eqref{eq:matrixAB}. 
\State 2) Compute the SVD of $\AAA$. 
\State 3) Compute   the columns of $\R$ using Remark \ref{rem:1}.
\State \textbf{output}: matrix $A_k^\star =\R\R^*\BBB V_\AAA\Sigma_\AAA^{-1}W_\AAA^*$
\end{algorithmic}
\caption{Solver for \eqref{eq:prob1} \label{algo:1}}
\end{algorithm}

\begin{algorithm}[t]
\begin{algorithmic}[0]
\State \textbf{input}: matrices $(\R, \Q,\theta)$, with $\Q= (\BBB\AAA^{\dagger})^* \R$
\State 1) Compute the SVD of matrix  $\Q$.
\State 2) Solve for $ i=1 \cdots k$ the eigen equation $ \tilde A_k w_i = \lambda_i w_i, $ where  $w_i \in \Cr^m$ and  $\lambda_i \in \Cr$ denote eigenvectors and eigenvalues of $\tilde  A_k=W_{\Q}^*\R V_{\Q}\Sigma_{\Q} \in \Rr^{m\times m}.$


\State \textbf{output}: DMD modes $  \phi_i=W_{\Q} w_i$ and amplitudes $\nu_{i,t} =  \lambda_i^{t-1} \phi_i^*\theta$
\end{algorithmic}
\caption{Low-rank DMD modes and amplitudes \label{algo:2}}
\end{algorithm}

\subsection{Reduced-Order Models}\label{sec:ROM}
We now discuss the resolution of  the reduced-order model \eqref{eq:model_koopman_approx} given the solution $A_k^\star$ of \eqref{eq:prob1}.
Trajectories of \eqref{eq:model_koopman_approx} are fully determined  by a $k$-dimensional recursion 
involving the projected variable  $z_t= \R^* \tilde x_t$:
\begin{equation}\label{eq:GaussLinSystReduced}
 \left\{\begin{aligned}
& z_t=\R^*\BBB \AAA^{\dagger}\R  z_{t-1} ,\\
&z_2=\R^*\BBB \AAA^{\dagger}\theta.  \\
\end{aligned}\right.
\end{equation}
Then, by multiplying both sides by matrix $\R$,   we  obtain the sought low-rank approximation $\tilde x_{t} = \R  z_t$.

Alternatively,  we can employ reduced-order model~\eqref{eq:koopman1}. The parameters of this model, \ie low-rank DMD modes and amplitudes, are efficiently computed without any minimization procedure, in contrast to what is proposed by the author in~{\cite{Jovanovic12}}. Indeed, we  rely on the following remark stating that  DMD modes and amplitudes can be obtained by means of SVDs using  Algorithm \ref{algo:2}. The remark is proved in  the  {technical report  \cite{Heas16_DMD}}.
 \begin{remark}\label{rem:2}
Each pair $(\phi_i,\lambda_i)$ generated by Algorithm \ref{algo:2}  is one of the $k$ eigenvector/eigenvalue pair of $A_k^\star$.
\end{remark}

\section{Numerical Evaluation}\label{sec:numEval}\vspace{-0.cm}
\begin{figure}[t!]
\centering\vspace{-3.95cm}
\begin{tabular}{c}
\includegraphics[width=0.9\columnwidth]{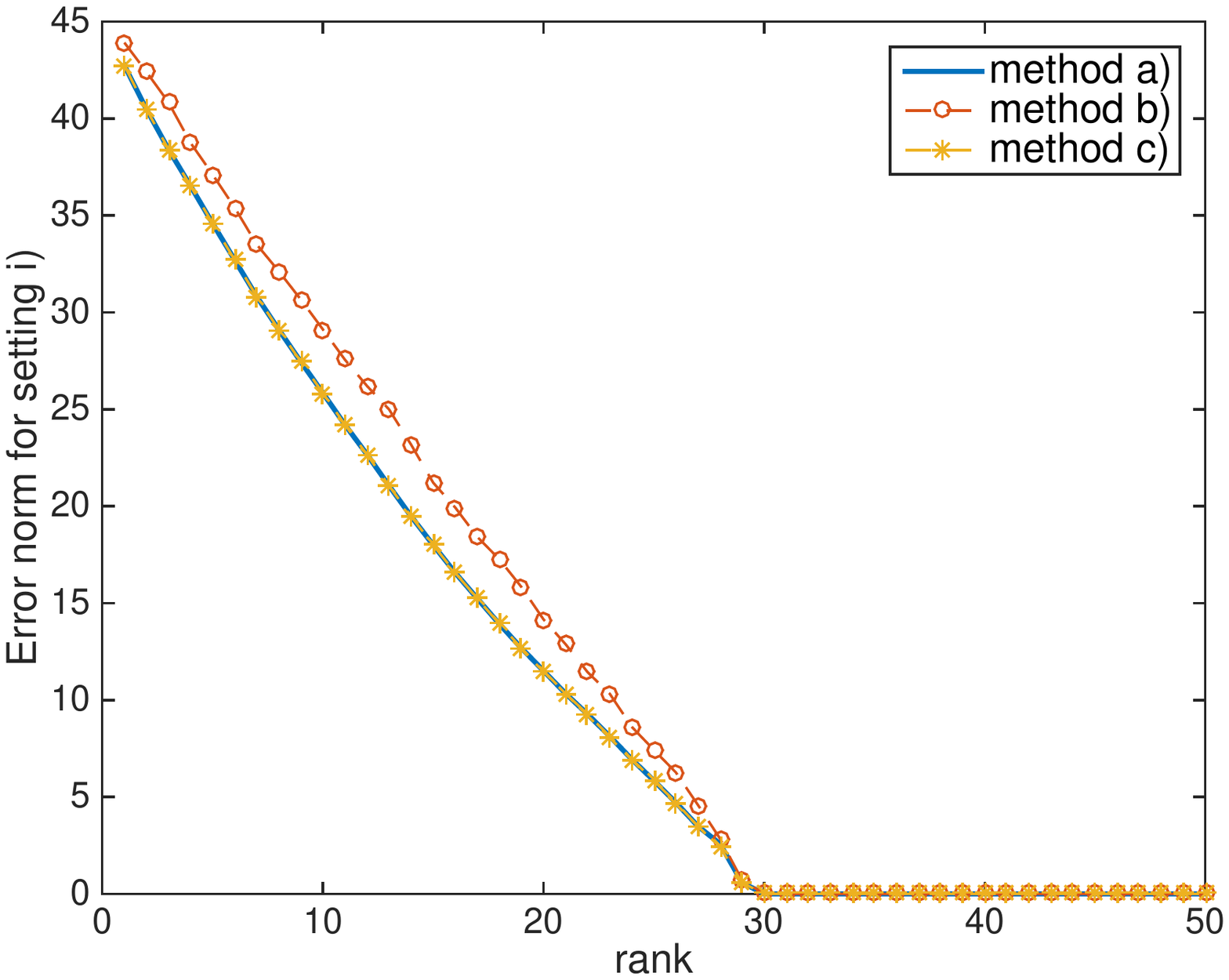}\vspace{-5.2cm}\\
\includegraphics[width=0.9\columnwidth]{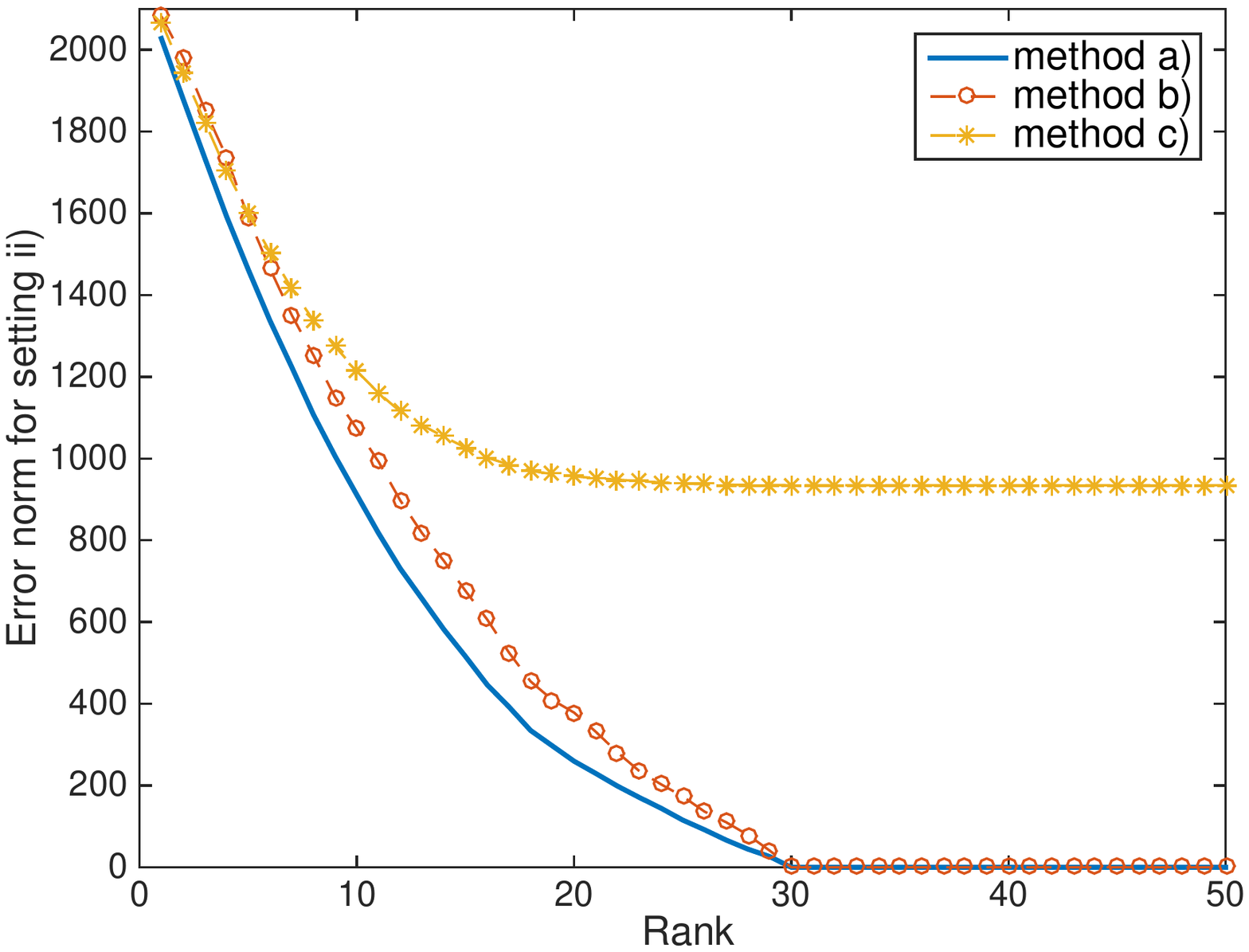}\vspace{-5.2cm}\\
\includegraphics[width=0.9\columnwidth]{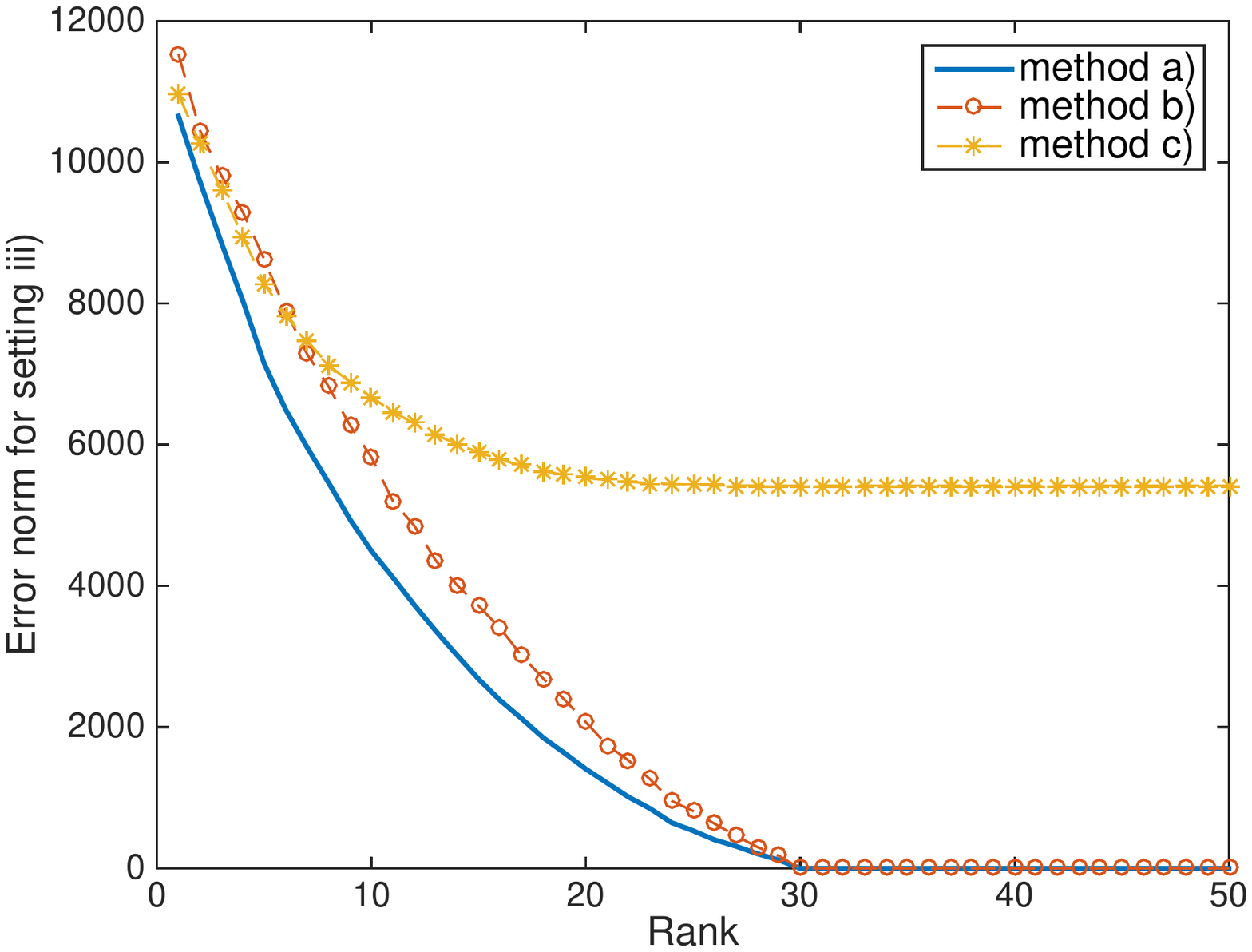}\vspace{-2.75cm}
\end{tabular}
\caption{\small Evaluation of error norms $\|\BBB-\hat A_k \AAA\|_F$   as a function of   rank~$k$.   Setting $i)$  {({top})} and $ii)$ (middle)  imply both a linear model but the former satisfies the snapshots linear dependence assumption. Setting $iii)$ {({bottom})} implements a non-linear model. We evaluate 3  algorithms:  {method $a)$} is the proposed optimal algorithm, {method $b)$} provides the rank-$k$ SVD approximation of the unconstrained solution given in \cite{Tu2014391} and {method $c)$} is the low-rank projected DMD method proposed in \cite{Jovanovic12}. See details in Section~\ref{sec:numEval}. \vspace{-0.3cm}\label{fig:1}}
\end{figure}

In what follows, we evaluate on a toy model the different  approaches for solving the low-rank DMD approximation problem. 
We consider a high-dimensional space of $n=50$ dimensions, a low-dimensional subspace of $r=30$ dimensions and  $m=40$ snapshots. Let $G$ 
be a matrix of rank $r$ generated randomly according to  $G=\sum_{i=1}^{r} \xi_i \xi_i^*$, 
where entries of $\xi_i$'s are $n$ independent samples of the standard normal distribution.   Let  the initial condition $\theta$ be randomly chosen according to the same distribution.  The snapshots, gathered in matrices $\AAA$ and $\BBB$, are generated using  \eqref{eq:model_init} 
for three  configurations of $f_t$:   \vspace{-0.2cm}
\begin{itemize}
\item[$i)$] $f_t(x_{t-1})=G x_{t-1}$, s.t. $\exists A^c$ satisfying $G \AAA=\AAA  A^c$,\vspace{-0.15cm}
\item[$ii)$]  $f_t(x_{t-1})=G x_{t-1}$,\vspace{-0.15cm}
\item[$iii)$]  $f_t(x_{t-1})=G x_{t-1}+G \textrm{diag}(x_{t-1})\textrm{diag}(x_{t-1})x_{t-1}$.\vspace{-0.15cm}
\end{itemize}
Setting $i)$  corresponds to a linear system satisfying the  assumption  \eqref{eq:companion}, as made in the projected DMD approaches \cite{Schmid10,Jovanovic12}. Setting $ii)$ and $iii)$, do not make this assumption and simulate respectively linear and non-linear dynamical systems.
We assess three different methods for computing $\hat A_k$: \vspace{-0.1cm}
\begin{itemize}
\item[$a)$]  optimal rank-$k$ approximation  given by Algorithm \ref{algo:1},\vspace{-0.1cm}
\item[$b)$] $k$th-order SVD approximation of  \eqref{eq:exactDMD}, \ie $k$-th order approximation of the rank-$m$ non-projected DMD solution  \cite{Tu2014391},\vspace{-0.1cm}
\item[$c)$]  rank-$k$ approximation by \eqref{eq:projDMD}, corresponding to the projected DMD approach \cite{Jovanovic12} (or \cite{Schmid10} for $k\ge m$).\vspace{-0.1cm}
\end{itemize}
The performance is measured in terms of the error norm $\|\BBB-\hat A_k \AAA\|_F$ with respect to  the rank~$k$. Results for the three settings are displayed in Figure \ref{fig:1}. 

As a  first remark, we notice  that the solution provided by Algorithm \ref{algo:1} (method $a$) yields the best results,  in agreement with  {T}heorem~\ref{prop22}. 

Second, in setting  $i)$, the experiments confirm  that when the linearity assumption is valid,  the low-rank projected DMD (method~$c$) achieves the same performance as the optimal solution (method~$a$). Moreover, truncating  the rank-$m$ DMD solution (method~$b$) induces as expected an  increase of the error norm. This deterioration is however moderate in our experiments. 

Then, in  settings $ii)$ and $iii)$ we remark that the behavior of the error norms are  analogous  (up to an order of magnitude).  The {performance} of the projected approach (method~$c$) differs notably from the optimal solution. A significant deterioration is visible for $k>10$. This is the consequence of the non-validity of the assumption made in method~$c$. Nevertheless, we notice that method~$c$ accomplishes a slight gain in performance compared to method~$b$ up to a moderate rank ($k<5$).  
Besides, we also notice that the error norm of method~$b$  in the case $k<30$ is not optimal. 
 
Finally, as expected, all methods succeed  in properly  characterizing the low-dimensional subspace as soon as $k \ge r$.

\section{Conclusion}
Following recent attempts to characterize an optimal low-rank approximation based on DMD, this paper provides a closed-form solution  to this non-convex optimization problem. To the best of our knowledge,  state-of-the-art methods are all  sub-optimal. The paper further proposes effective algorithms based on SVD to solve this problem and run  reduced-order models. 
Our numerical experiments attest that the proposed algorithm is more accurate than state-of-the-art methods. In particular, we illustrate the fact that  simply truncating the full-rank DMD solution, or  exploiting too restrictive assumptions for the approximation subspace is insufficient.   
 
\newpage 

\section*{Acknowledgements}
This work was
supported by the  ``Agence Nationale de la Recherche" through the GERONIMO project (ANR-13-JS03-0002).

\appendix
\bibliographystyle{IEEEbib}
\bibliography{./bibtex}

\end{document}